\documentclass[conference]{IEEEtran}
\IEEEoverridecommandlockouts

\usepackage{graphicx}
\usepackage{amsmath}
\usepackage{amssymb}
\usepackage{booktabs}
\def\eg{{\emph{e.g.}}}
\def\ie{{\emph{i.e.}}}

\def\ie{\emph{i.e.}}
\def\eg{\emph{e.g.}}
\def\x{{\mathbf{x}}}
\def\z{{\mathbf z}}

\def\eg{{\emph{e.g.}}}
\def\ie{{\emph{i.e.}}}

\def\s{{\mathbf{s}}}

\def\M{{\mathbf{M}}}

\def\G{{\mathbf{\mathcal{G}}}}

\def\s{{\mathbf{s}}}

\def\G{{\mathcal{G}}}
\def\M{\mathbf{M}}

\newcommand{\Tref}[1]{Table~\ref{#1}}
\newcommand{\Eref}[1]{Eq.~\ref{#1}}
\newcommand{\Fref}[1]{Fig.~\ref{#1}}

\usepackage[pagebackref,breaklinks,colorlinks]{hyperref}

\usepackage[capitalize]{cleveref}
\crefname{section}{Sec.}{Secs.}
\Crefname{section}{Section}{Sections}
\Crefname{table}{Table}{Tables}
\crefname{table}{Tab.}{Tabs.}
\newcommand{\renjie}[1]{\textcolor{black}{{#1}}}

\newcommand{\yf}[1]{\textcolor{black}{{#1}}}
\newcommand{\yufei}[1]{\textcolor{black}{{#1}}}

\begin{document}

\title{Removing Image Artifacts From Scratched Lens Protectors}

\author{\IEEEauthorblockN{Yufei Wang\textsuperscript{\rm1}, Renjie Wan\textsuperscript{\rm2}, Wenhan Yang\textsuperscript{\rm3}, Bihan Wen\textsuperscript{\rm1*}\thanks{*: Corresponding author.}, Lap-Pui Chau\textsuperscript{\rm4}, Alex C. Kot\textsuperscript{\rm1}}
\IEEEauthorblockA{
\textsuperscript{\rm1} School of Electrical and Electronic Engineering, Nanyang Technological University, Singapore\\
\textsuperscript{\rm2} Department of Computer Science, Hong Kong Baptist University, Hong Kong\\
\textsuperscript{\rm3} Distributed High Performance Software Fundamental Research Laboratory, Pengcheng National Lab, China\\
\textsuperscript{\rm4} Department of Electronic and Information Engineering, Hong Kong Polytechnic University, Hong Kong\\
\{yufei001, bihan.wen, eackot\}@ntu.edu.sg, renjiewan@comp.hkbu.edu.hk, yangwh@pcl.ac.cn, lap-pui.chau@polyu.edu.hk}}

\maketitle

\begin{abstract}
    \renjie{A protector is placed in front of the camera lens for mobile devices to avoid damage, while the protector itself can be easily scratched accidentally, especially for plastic ones. The artifacts appear in a wide variety of patterns, making it difficult to see through them clearly. Removing image artifacts from the scratched lens protector is inherently challenging due to the occasional flare artifacts and the co-occurring interference within mixed artifacts. \yufei{Though different methods have been proposed for some specific distortions}, they seldom consider such inherent challenges. In our work, we consider the inherent challenges in a unified framework with two cooperative modules, which facilitate the performance boost of each other.  We also collect a new dataset from the real world to facilitate training and evaluation purposes. The experimental results demonstrate that our method outperforms the baselines qualitatively and quantitatively. The code and datasets will be released at \url{https://github.com/wyf0912/flare-removal}}
\end{abstract}

\section{Introduction}
\label{sec:introduction}
\renjie{
A rising number of works aims to improve the visibility of images \cite{wang2022low, wan2022benchmarking, wan2022purifying, wan2021face}, while they usually assume that the device itself is intact.}
{However, the protector of the lens can be easily scratched accidentally, especially low-cost devices, leading to scratch-related artifacts in the captured image. These artifacts can reduce image contrast and cause errors for the follow-up tasks (\eg, classification~\cite{wang2020heterogeneous, wang2022variational}, and segmentation~\cite{wang2021embracing}). Thus, how to suppress these artifacts becomes a meaningful task with practical values to investigate.}



\renjie{Scratch-related artifacts are caused by unwanted light reflections and scattering in an optical system. While the lens is supposed to be in focus, the scratches easily deviate the incoming light rays~\cite{goro1953}. Such deviations lead to flare artifacts for light rays emitted by intense light source~\cite{lee2013practical}, which occludes the desired information during the post-processing, \eg, dynamic range compression~\cite{fattal2002gradient}. Besides, for light rays without sufficient energy, such deviations lead to haze-like artifacts within a specific area, where the scene is degraded by several distortions (\eg, blurring, scattering, and attenuation). Generally, due to the vulnerability of plastic-made protectors, scratch-related artifacts are more serious than other camera-related artifacts (\eg, dirty lens) discussed in previous methods~\cite{li2021let, gu2009removing, Zamir2021MPRNet, li2021you}, which increases its challenge to solve this problem in a usual way.}

\begin{figure}[t]
    \centering
    \includegraphics[width=1\linewidth, clip, trim=80 10 30 0]{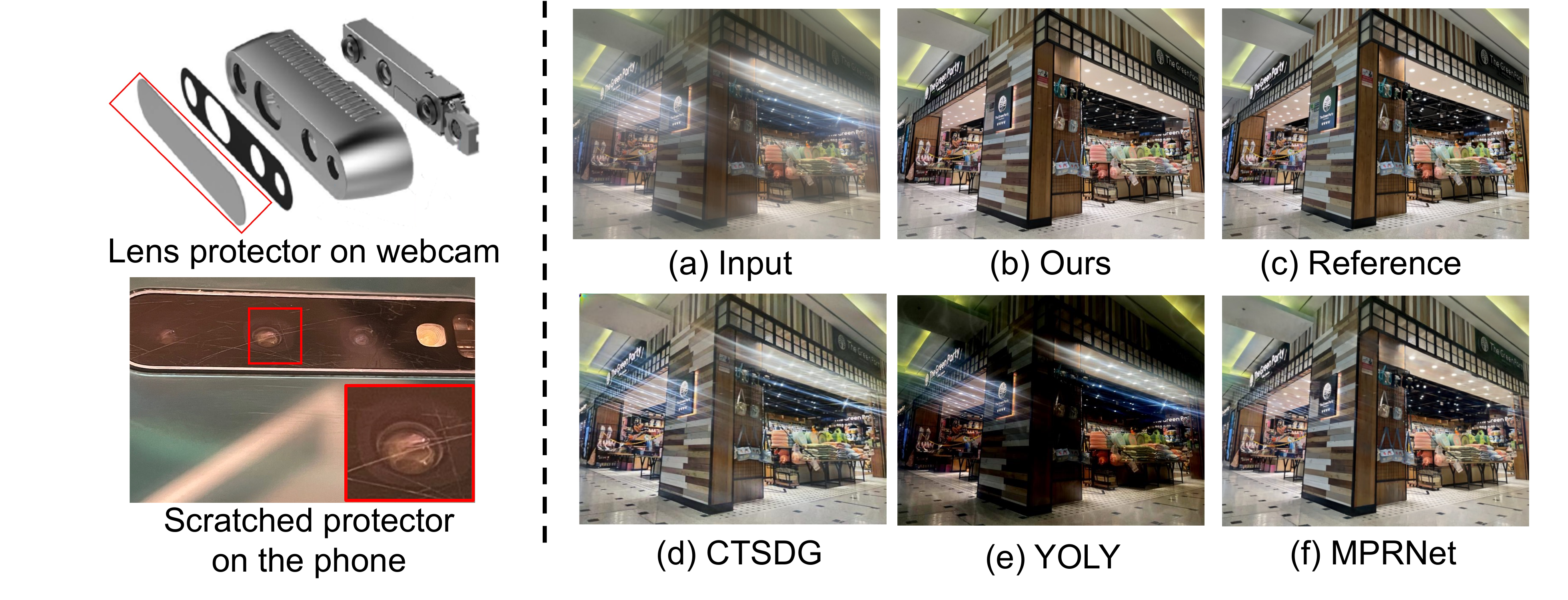}
    \caption{\yf{Left part: The lens protector on the webcam and scratched protector on a mobile phone. Right part: (a) The degraded image with scratched related artifacts, (c) its reference image, and the results generated by (b) our method , (d) CTSDG~\cite{Guo_2021_ICCV}, (e) YOLY~\cite{li2021you}, and (f) MPRNet~\cite{Zamir2021MPRNet}}.}
    \label{fig:firstimage}
    \vspace{-0.5cm}
\end{figure}

\renjie{The first challenge comes from the co-occurring interference within diverse artifacts in the optimization process. Due to the irregular property of the scratch \cite{wu2021train, asha2019auto, Floris2015, vitoria2019automatic}, the related artifacts present various properties (degradation levels, contrast, and positions). To recover the desired light rays deviated by the scratch, the models need to simultaneously handle such various artifacts from a single observation. However, the purely focus on one single type of artifact may lead to the performance drop on other artifacts. Besides, the various artifacts both get involved in modeling each other during image formation, which further increases the ill-posedness of this problem. Without considering such co-occurring interference, the results obtained by naively training an image restoration networks (\eg, MPRNet~\cite{Zamir2021MPRNet} and YOLY~\cite{li2021you} in~\Fref{fig:firstimage}) show noticeable residue artifacts, though the scratch-related artifacts look similar to the degradation they target on.}

\renjie{Another challenge comes from how to handle the occasional flare artifacts. The flare artifacts only occasionally appear in the captured image. However, it may seriously occlude the desired information and influence the image's overall perception. Though several methods have been proposed to handle flare artifacts~\cite{wu2021train} or other similar occlusion effects (\eg, image inapinting~\cite{Guo_2021_ICCV, song2018contextual, dolhansky2018eye,iizuka2017globally,pathak2016context}, the raindrops on the glass~\cite{quan2019deep, eigen2013restoring}), they always tend to train a network based on a large number of synthetic images with such artifacts.}  Besides, due to the ignorance of the co-occurring interference, it makes those occlusion-oriented methods difficult to handle other non-flare artifacts.

\begin{figure*}[tbp]
    \centering
    \includegraphics[width=1\textwidth, trim=10 10 10 10,clip]{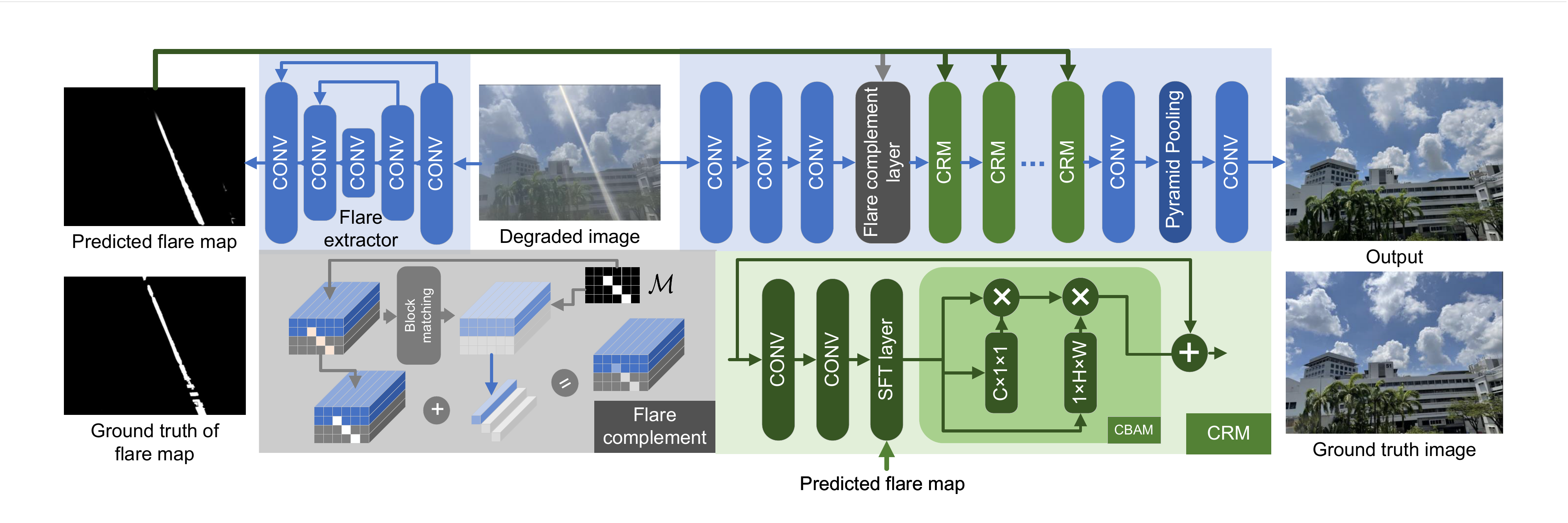}
    \vspace{-0.3cm}
    \caption{\renjie{The overall framework \renjie{of our proposed approach}. We employ a flare attention module to tackle the missing information occluded by flare artifacts and a co-occurring refinement module (CRM) to tackle co-occurring interference based on channel and spatial-wise attentions. The two modules are deployed into one network to better learn the properties of co-occurring interference by sharing the information flow. A flare extractor is additionally used to localize the flare maps.}}
    \label{fig:framework}
\end{figure*}

\renjie{In this work, to address the above challenges, we discuss the properties of scratch-related artifacts and propose to handle them in a unified but differentiated framework. As shown in~\Fref{fig:framework}, for the differentiation between flare and other artifacts, we employ a flare attention module to tackle the occluded information by utilizing the internal similarity. Specifically, we propose to introduce a co-occurring refinement module to consider the flare artifacts and other artifacts in a unified module to address the co-occurring interference. Besides, we directly deploy the two modules into one network, where they can boost the performance of each other and better learn the property of co-occurring interference by sharing the information flow. At last, we introduce an operable strategy to collect real data for training and evaluation purposes.}

\section{Proposed method}
\noindent\textbf{{Dataset collection.}}
We introduce a scratch image set $(\x_d, \x_{gt})$ for training and evaluation, where $\x_d$ and $\x_{gt}$ denote the image captured with the scratch and normal protector, respectively. 
For each device, we prepare two external protectors (one for normal and one for scratch) with similar properties to the original protector. The external protectors are all randomly scratched to cover more diversified degradation. Then, to avoid the additional scattering and reflection effects \yf{between the original lens protector and our added one}, we remove the original protector and capture images using a two-steps strategy as follows: 1) Taking the normal image $\x_{gt}$ with the normal protector; 2) capturing the scratched image $\x_d$ with the scratch protector. Finally, we capture $1251$ sets of images under diversified scenarios (\eg, indoor, outdoor, and lighting conditions). 281 sets among them are used for the evaluation, and 970 sets are used for training purposes.


\noindent\textbf{{Flare maps.}} \renjie{To facilitate the investigation of flare artifacts, we further annotate flare regions for images with flare artifacts.} \renjie{Despite the labor-intensive manual annotations, we propose to first subtract the area mean illumination as} 
\begin{equation}
    l(\x) = \mu(\x)-\psi(\mu(\x)),
\end{equation}
\renjie{where $\mu(\cdot)$ converts the original image into its grayscale version to attenuate the side effect from color distortion and $\psi(\cdot)$ denotes the average pooling operation used to suppress the reduction of the image contrast. Then, the ground truth of flare regions can be obtained as follows:}
\begin{equation}
    f(\x_{d}, \x_{gt}) = (T((l(\x_d) - l(\x_{gt}))\odot \mu(\x_d)) \ominus B) \oplus B,
    \label{eq:flare_gt}
\end{equation}
\renjie{where $\odot$ is the element-wise multiplication, $\ominus$ and $\oplus$ denote the erosion and dilation operations, respectively, and $B$ denotes the structuring element for morphological operation. Specifically, $T(\cdot)$ is a binary operation defined as}
\begin{equation}
	T(x)= \left\{\begin{matrix} 0, & x \leq 0.002, \\
		1, & \text{otherwise}.  \\
	\end{matrix}\right.
\end{equation}

To better differentiate different artifacts in our unified framework, we use a flare extractor based on U-Net~\cite{ronneberger2015u} to localize the flare regions as follows:
\begin{equation}
	\M = \G_{flare}(\x_{d}),
	\label{eq:l_flare}
\end{equation}
\renjie{where $\G_{flare}$ denotes the flare extractor used to localize flare artifacts, and $\M$ denotes the estimated flare maps.}	

\noindent\textbf{{Flare attention module.}}
We propose to predict missing pixels in the flare area $\M$ which is obtained in~\Eref{eq:l_flare} using the information from the image itself. The formulation is as follows
\begin{equation}
\small
	\s_{x,y,x',y'} = <\frac{\z_{x,y}}{||\z_{x,y}||},\frac{\z_{x',y'}}{||\z_{x',y'}||}>,
	\label{eq:context_attention}
\end{equation}
where $\s_{x,y,x',y'}\in \mathbb{R}^{W\times H\times W\times H}$ denotes the similarity between feature vectors $\z_{x, y}$ and $\z_{x', y'}$.

Based on~\Eref{eq:context_attention}, we do similar steps with~\cite{yu2018generative} that reshape the similarities $\s'$ to $\mathbb{R}^{WH\times W\times H}$ and conduct softmax among $x',y'$ dimension as follows
\begin{equation}
	\mathbf{w} = \text{softmax}_{(\cdot,x',y')} (\text{reshape}(\s)),
\end{equation}
where $\mathbf{w}$ is the output weight with the dimension $\mathbb{R}^{WH\times W\times H}$. Finally, the output of our flare attention module is as follows
\begin{equation}
\begin{split}
\z' = \text{Deconv}(\mathbf{w},\z) \odot \mathcal{M} + \z \odot (1-\mathcal{M}),
\end{split}
\end{equation}
where $\text{Deconv}(\mathbf{w},\z)$ represents the deconvolution operation that we take $\mathbf{w}\in \mathbb{R}^{WH\times W\times H}$ as input and use $\z$ as the weight of deconvolution kernel by reshaping it to $\mathbb{R}^{WH\times C \times 1\times  1}$.



\noindent\textbf{Co-occurring refinement module.}
To better suppress the co-occurring interference with diverse properties, we propose the co-occurring refinement module (CRM) to consider the haze-like artifacts removal and flare features refinement in one module. For haze-like artifacts, by embedding the Convolutional Block Attention Module (CBAM)~\cite{woo2018cbam} into the classical vanilla residual block. The \yf{combined channel and spatial attention modules} of CBAM facilitates the exploration of the global contextual information across channels and extraction of the inter-spatial feature relationship \yf{simultaneously}. Specifically, for an input feature $\z\in \mathbb{R} ^{C\times H \times W}$, the output features of CBAM can be defined as follows:
\begin{equation}
\begin{split}
    \z' &= \tau_c(\z) \cdot \z, \\
    \z'' &= \tau_s(\z') \cdot \z',
\end{split}
\end{equation}
where $\tau_c(\z) \in \mathbb{R} ^{C\times 1 \times 1}$ and $\tau_s(\z') \in \mathbb{R} ^{1\times H \times W}$ are the channel attention map and spatial attention map defined as follows:
\begin{equation}
\small
\begin{split}
     \tau_c(\z) &= \sigma( MLP(AvgPool(\z))+MLP(MaxPool(\z)) ), \\
     \tau_s(\z) &= \sigma( conv^{7\times7}([\psi(\z);MaxPool^{7\times7}(\z)])),
\end{split}
\end{equation}
where $\sigma$ is the sigmoid function, $AvgPool$ and $MaxPool$ are global average pooling and global maximum pooling that return features with the size $\mathbb{R} ^{C\times 1 \times 1}$, and $\psi(\z)$ and $MaxPool^{7\times7}(\z)$ are average pooling and global pooling that return the feature with the same size as the input. The channel number of the latent layer in  multi-layer perceptron (MLP) is set to $C/16$ to reduce parameter overhead.

\renjie{To further refine the features from the flare attention module, we propose to embed the Spatial Feature Transform (SFT) layer into our CRM module as guidance. Specifically, we model the flare maps as the parameters of affine transformation in SFT as follows:}
\begin{equation}
    \z' = \gamma (\M) \cdot \z + \beta (\M),
\end{equation}
\renjie{where $\z$ denotes the input features of the SFT layer~\cite{wang2018recovering}, $\z'$ is the output of the SFT layer, and  $\gamma$ and $\beta$ are two networks with two convolutional layers for the prediction of the scale and bias, respectively. \yf{For the image without flare artifacts, \ie{, $\mathbf{M}$ is all zero}, $\gamma (\M)$ and $\beta (\M)$ are two spatial invariant maps.} Specifically, spatial pyramid pooling~\cite{he2015spatial} is also employed after our CRM to better aggregate features from different scales. Finally, the features from the spatial pyramid pooling layer are concatenated and then fed into a convolutional layer to obtain a clean image.}

\noindent\textbf{Loss functions}
The whole loss functions is defined as follows:
\begin{equation}
\small
\mathcal{L}_G= \alpha \mathcal{L}_{1} + \beta \mathcal{L}_{flare}  + \omega\mathcal{L}_{adv}^G,
\end{equation}
where $\alpha=1$, $\beta=1$, and $\omega=0.002$ are the weighting coefficients for each loss, and $L_{flare}$ and the adversarial loss $\mathcal{L}_{adv}$ are defined as follows,
\begin{equation}
\small
\begin{split}
    &\mathcal{L}_{flare} = ||\mathbf{M} - \mathbf{M}^{*}||_2, \\
    &\mathcal{L}_{adv}^G = -\log(D(x_{gt}, G(x_d)))-log(1-D(G(x_d),x_{gt})),
\end{split}
\end{equation}
where $D$ is the discriminator.

\section{Experiments}
\renjie{Due to the lack of directly related methods, we instead compare the performance of our method with other image restoration methods. Considering the blurring effects and haze effects caused by scratch-related artifacts, we first compare with the state-of-the-art methods for image deblurring~\cite{Zamir2021MPRNet} and image dehazing~\cite{li2021you}. Then, we also compare with CycleGAN~\cite{CycleGAN2017} and Pixel2Pixel~\cite{isola2017image}, two classical image translation methods. Besides, to evaluate the effectiveness of our flare removal ability, we compare it with an image inpainting method CTSDG~\cite{Guo_2021_ICCV}. For dirty camera setting, we consider DISCNet~\cite{Feng_2021_CVPR} which is designed to remove the diffraction artifacts from under-display cameras. Since those methods are all not designed for our problem, for a fair comparison, we train their models on our collected dataset so that they can map the input degraded image to an output degradation-free image, as in our model.}{We adopt LIPIS~\cite{zhang2018perceptual} as the error metric which measures perceptual image similarity. The lower LPIPS values indicate better performances. We also choose SSIM and PSNR metrics to do the comparisons.}

\begin{figure*}[tbp]
    \centering
    \includegraphics[width=1\linewidth, trim=0 250 0 320,clip]{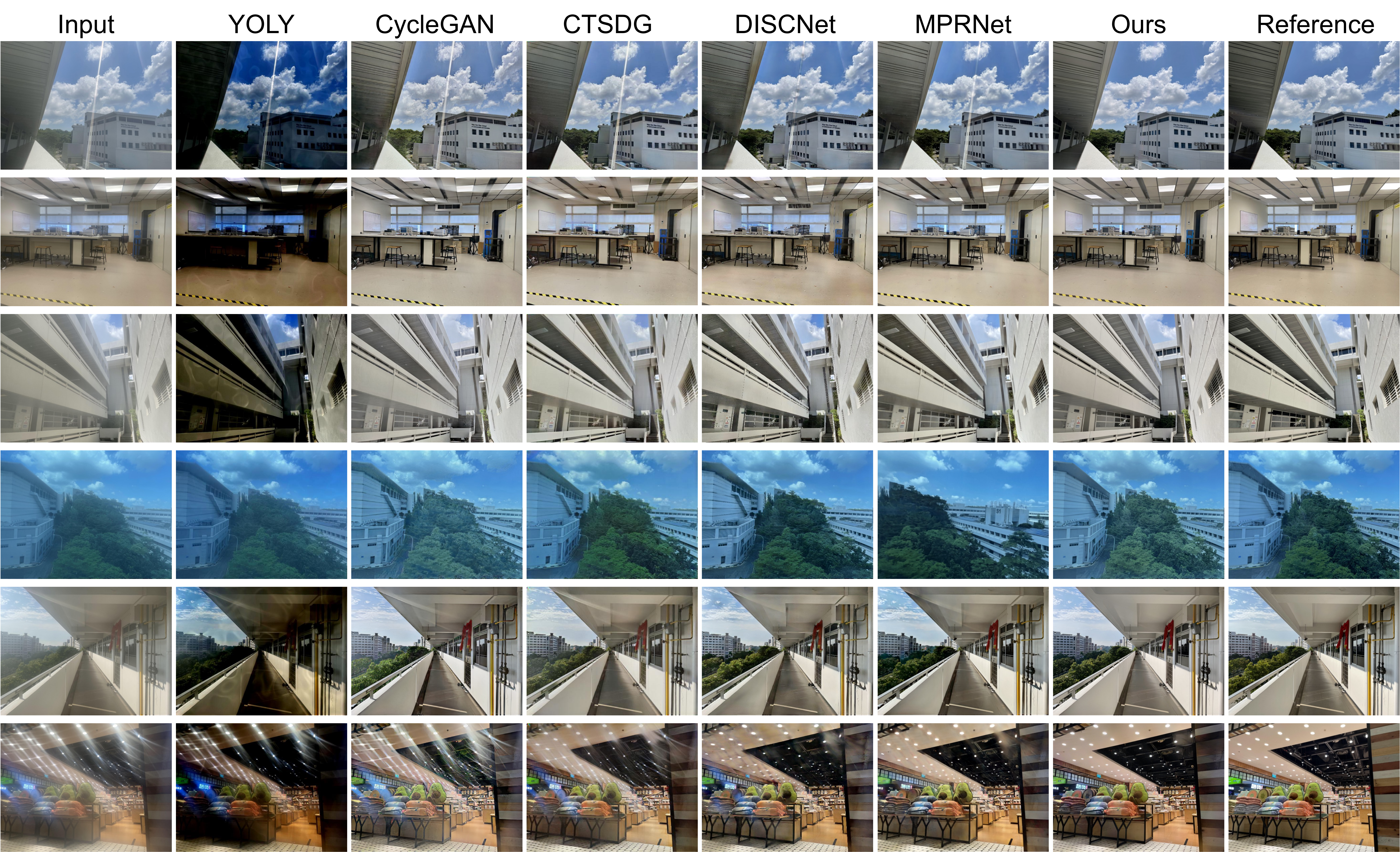}
    \vspace{-0.3cm}
    \caption{Examples of results on our collected dataset. From left to right, the input, the results obtained by YOLY~\cite{li2021you}, CycleGAN~\cite{CycleGAN2017}, CTSDG~\cite{Guo_2021_ICCV}, DISCNet~\cite{Feng_2021_CVPR}, MPRNet~\cite{Zamir2021MPRNet}, and our method, and the reference image. Please zoom in to see the details.}
    \vspace{-4mm}
    

    \label{fig:results}
\end{figure*}

\subsection{Experimental results}
\noindent\textbf{Quantitative evaluations.} We evaluate the performance of our method and other competitors \renjie{on our collected evaluation dataset. From the quantitative results shown in~\Tref{tab:result}, our method achieves more promising performance than other competitors. The higher LPIPS values demonstrate that our proposed method achieves better visually pleasing results and the higher PSNR values indicate better color consistency. Besides, the results in~\Tref{tab:result} also indicate that the supervised methods using paired data achieve generally better performance than other weakly-supervised methods (\eg, CycleGAN~\cite{CycleGAN2017}). It partly proves the supervised models' effectiveness in handling such tasks and the necessity of our collected paired data. Besides, as a method designed for image restoration across a range of tasks, MPRNet~\cite{Zamir2021MPRNet} still cannot outperform our methods either with its pretrained model for deblurring tasks or the model retrained on our dataset.} 




\noindent\textbf{Qualitative evaluations.} \renjie{The qualitative comparisons are shown in~\Fref{fig:results}. Our method largely enhances the visibility of target scenes by suppressing scratch-related artifacts. For example, from the results in the first and last row of~\Fref{fig:results}, our method effectively suppresses the flare artifacts and successfully recovers the occluded information, while almost all other competitors cannot effectively handle the flare artifacts including the inpainting based method CTSDG~\cite{Guo_2021_ICCV}. Besides, for other types of artifacts with a similar appearance to blur and haze, our method still shows superior performance compared with YOLY~\cite{li2021you} and MPRNet~\cite{Zamir2021MPRNet}, two methods specifically designed for dehazing and deblurring, respectively.}


\begin{table}[htbp]
    \centering
        \caption{Quantitative evaluations on our collected dataset using three different metrics, and compared with CycleGAN~\cite{CycleGAN2017}, Pixel2Pixel \cite{isola2017image}, DISCNet~\cite{Feng_2021_CVPR}, MPRNet$^*$~\cite{Zamir2021MPRNet}, MPRNet~\cite{Zamir2021MPRNet}, YOLY \cite{li2021you}, and CTSDG \cite{Guo_2021_ICCV}. MPRNet$^*$ represents the results that directly using the released pre-trained model for deblurring.}  
    \scalebox{1}{
    \vspace{-0.2cm}
    \begin{tabular}{ccccc}
    \toprule
         & PSNR$\uparrow$ & SSIM$\uparrow$ & LPIPS$\downarrow$ \\
    \midrule
    Input & 17.94 & 0.6625 & 0.3232 \\
     CylceGAN \cite{CycleGAN2017}& 17.49 & 0.6442 & 0.4249\\
     Pixel2Pixel \cite{isola2017image}& 19.14 & 0.7897 & 0.1793\\
     DISCNet \cite{Feng_2021_CVPR}& 20.78 & 0.7528 & 0.2887 \\
     MPRNet$^*$ \cite{Zamir2021MPRNet} & 16.74 & 0.6259 & 0.4624 \\
     MPRNet \cite{Zamir2021MPRNet} & 20.04 & 0.7037 & 0.3549\\
     YOLY \cite{li2021you} & 12.01 & 0.4608 & 0.5088 \\
     CTSDG \cite{Guo_2021_ICCV} & 22.17 & 0.8176 & 0.2201 \\
     Ours & \textbf{23.43} & \textbf{0.8640} & \textbf{0.0927}\\
    \bottomrule
    \end{tabular}}
    \label{tab:result}
    \vspace{-0.3cm}
\end{table}

\begin{figure}[htbp]
    \centering
    \includegraphics[width=1\linewidth,clip,trim=0 20 0 0 ]{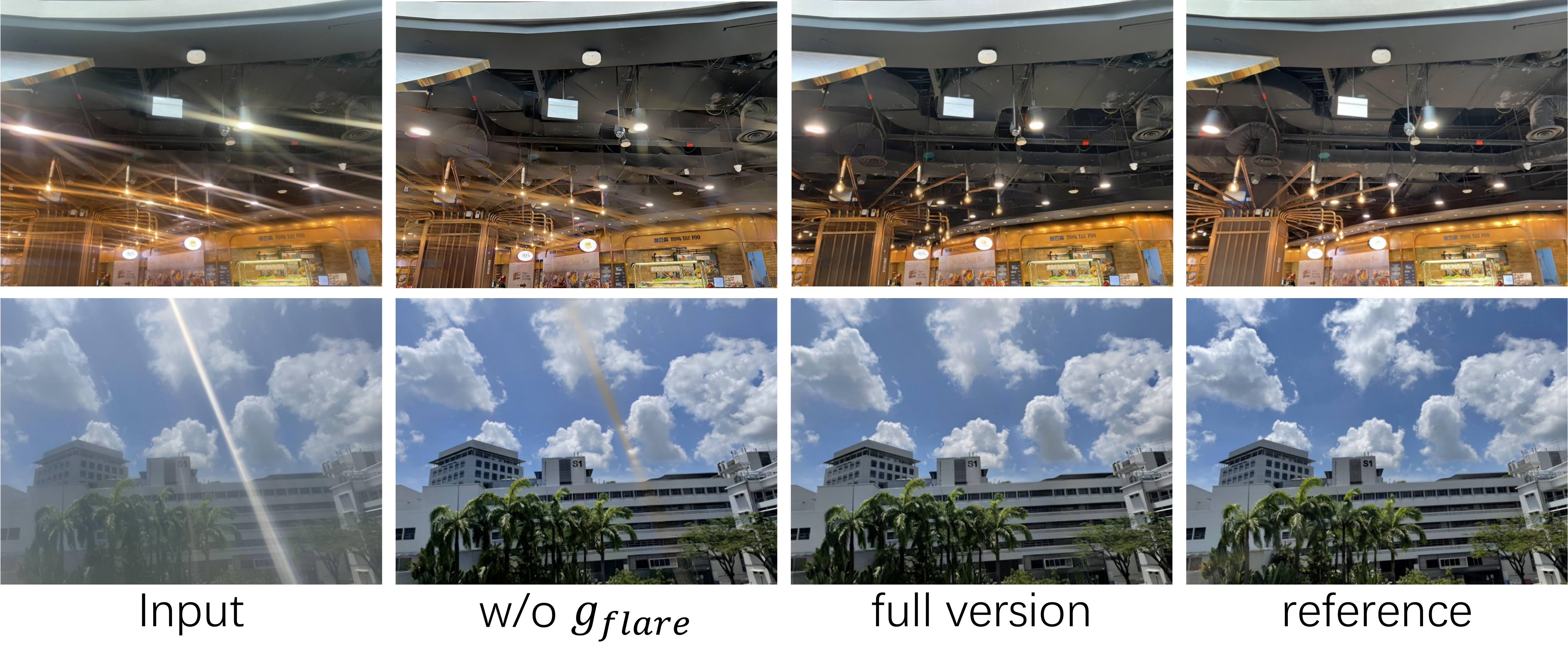}
    \vspace{-0.5cm}
    \caption{A comparison with and without the flare extractor $\G_{flare}$. The generated images by the model without $\G_{flare}$ tend to remain artifacts in severely damaged areas.}
    \label{fig:flare_extractor}
    \vspace{-0.5cm}
\end{figure}

\begin{table}[htbp]
    \centering
    \caption{\renjie{Quantitative evaluations for the model w/o the flare extractor $\G_{flare}$, the model w/o CBAM layers, the model w/o contextual attention module, and the complete model.}}
    \scalebox{1}{
    \begin{tabular}{cccc}
    \toprule
         & PSNR$\uparrow$ & SSIM$\uparrow$ & LPIPS$\downarrow$ \\
    \midrule
         w/o flare extractor $\mathcal{G}_{flare}$ & 22.17 & 0.8513 & 0.1246 \\
         w/o CBAM layers & 22.24 & 0.8289 & 0.1207 \\
         w/o flare attention module & 23.26 & 0.8561 & 0.0959\\
         Complete model & \textbf{23.43} & \textbf{0.8640} & \textbf{0.0927}\\
    \bottomrule
    \end{tabular}}
    \label{tab:ablation}
\end{table}

\subsection{Ablation study}
We conduct experiments to explore the effectiveness of network modules. As we can see in Table \ref{tab:ablation}, our complete model achieves the best performance which demonstrates the effectiveness of our proposed attention guided modules. We also compare the difference in the perceptual quality of the generated images.
For example, the comparisons w/ and w/o the contextual attention module are shown in Fig. \ref{fig:flare_extractor}. As we can see, though w/ or w/o $\G_{flare}$ can both achieve accurate color, the generated images without the guidance of $\G_{flare}$ tend to fail in recovering the image with strong artifacts.

\section{Conclusions}
We propose to remove scratch-related artifacts in this work. We first analyze the characteristic of scratch-related artifacts and collect a new dataset from the real world for training and evaluation purposes. Considering the inherent co-occurring interference within various scratch-related artifacts, we introduce a unified framework with the emphasis on flare artifacts and other non-flare artifacts, respectively. Our experiments have shown promising results for this problem.

\section*{Acknowledgment}
This work was carried out at the Rapid-Rich Object Search (ROSE) Lab, Nanyang Technological University (NTU), Singapore. The research is supported in part by the NTU-PKU Joint Research Institute (a collaboration between the NTU and Peking University that is sponsored by a donation from the Ng Teng Fong Charitable Foundation) and in part by the Start-Up Grant and NTU-Imperial Collaboration Fund (INCF-2022-003). Renjie Wan is supported by the Blue Sky Research Fund of HKBU under Grant No. BSRF/21-22/16 and Guangdong Basic and Applied Basic Research Foundation under Grant No. 2022A1515110692.

\newpage
{\small
\bibliographystyle{ieee_fullname}
\bibliography{PaperForReview}
}

\end{document}